\documentclass[letterpaper, 10 pt, conference]{ieeeconf} 

\IEEEoverridecommandlockouts

\usepackage{bm}
\usepackage{mathrsfs}
\usepackage{float}
\usepackage{caption}
\usepackage{tabularx}

\usepackage{cite}
\usepackage{amsmath,amssymb,amsfonts}
\usepackage{graphicx}
\usepackage{textcomp}
\usepackage{xcolor}
\usepackage{subfigure}
\usepackage{multirow}
\usepackage{booktabs}
\usepackage{hyperref}
\usepackage{marvosym}
\usepackage{algorithmicx}
\usepackage{algpseudocode}
\usepackage[ruled,linesnumbered]{algorithm2e}

\title{\LARGE \bf Ocean Diviner: A Diffusion-Augmented Reinforcement Learning Framework for AUV Robust Control in Underwater Tasks
}

\author{Jingzehua Xu$^\dagger,$\textsuperscript{\Letter}, Guanwen Xie$^\dagger$, Weiyi Liu$^\dagger$, Jiwei Tang, Ziteng Yang,\\Tianxiang Xing, Yiyuan Yang, Shuai Zhang and Xiaofan Li
\thanks{$\dagger$ These authors contribute to this work equally.}
\thanks{\textsuperscript{\Letter} Corresponding author. (Email: xjzh23@mails.tsinghua.edu.cn)}
}

\begin{document}

\maketitle
\thispagestyle{empty}
\pagestyle{empty}

\begin{abstract}
 Autonomous Underwater Vehicles (AUVs) are essential for marine exploration, yet their control remains highly challenging due to nonlinear dynamics and uncertain environmental disturbances. This paper presents a diffusion-augmented Reinforcement Learning (RL) framework for robust AUV control, aiming to improve AUV's adaptability in dynamic underwater environments. The proposed framework integrates two core innovations: (1) A diffusion-based action generation framework that produces physically feasible and high-quality actions, enhanced by a high-dimensional state encoding mechanism combining current observations with historical states and actions through a novel diffusion U-Net architecture, significantly improving long-horizon planning capacity for robust control. (2) A sample-efficient hybrid learning architecture that synergizes diffusion-guided exploration with RL policy optimization, where the diffusion model generates diverse candidate actions and the RL critic selects the optimal action, achieving higher exploration efficiency and policy stability in dynamic underwater environments. Extensive simulation experiments validate the framework’s superior robustness and flexibility, outperforming conventional control methods in challenging marine conditions, offering enhanced adaptability and reliability for AUV operations in underwater tasks. Finally, we will release the code publicly soon to support future research in this area.
    \end{abstract}

\vspace{-0mm}
\section{INTRODUCTION}
\label{sec:intro}
Autonomous Underwater Vehicles (AUVs) play a vital role in marine exploration, supporting missions from deep-sea mapping to infrastructure inspection \cite{1,2,3}. However, their performance in extreme ocean conditions faces three key challenges: (1) complex nonlinear hydrodynamics and thruster dynamics \cite{4}, (2) time-varying disturbances like currents and waves \cite{5}, and (3) localization uncertainty due to sensor limitations \cite{6}. These factors create a demanding control problem requiring simultaneous optimization of trajectory tracking, energy efficiency, and collision avoidance - often with conflicting objectives \cite{29}.

Traditional control approaches have made incremental progress in addressing these challenges. While PID controllers offer simplicity, their fixed-gain architecture cannot adapt to the wide range of hydrodynamic conditions encountered during typical missions \cite{7,23}. Model predictive control (MPC) provides better performance through optimization-based planning, but its computational complexity grows exponentially with prediction horizons, making real-time implementation impractical for resource-constrained AUVs \cite{8,24}. Sliding mode control (SMC) demonstrates robustness to bounded disturbances, but suffers from chattering effects that increase mechanical wear and energy consumption \cite{9,25}. Most critically, these methods lack the capacity to learn from operational experience or adapt to complex mission and environmental requirements.

\begin{figure}[!t]
        \centering
        \includegraphics[width=0.99\linewidth]{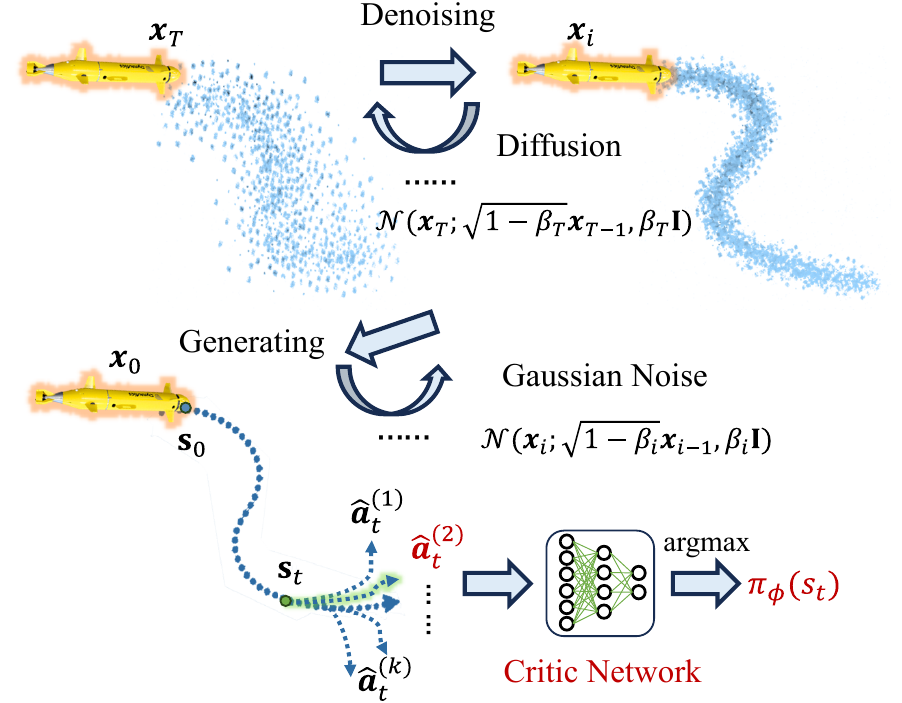}
        \caption{\small \textbf{Illustration of AUV control using the proposed framework}. This framework utilizes a diffusion model for action generation, while RL's critic network selects the action that gains the maximum Q value from candidates as the optimal action to execute.}
        \label{fig_1}\vspace{-1.8mm}
        \end{figure}

The advent of reinforcement learning (RL) has introduced new possibilities for robust AUV control \cite{28}. Through interacting with the environment, RL algorithms can learn sophisticated control policies \cite{10}. Recent work has demonstrated promising results in applications such as path following \cite{11} and station keeping \cite{12}. However, conventional RL approaches face two fundamental limitations in the AUV domain: (1) the sample inefficiency of exploration in high-dimensional state spaces \cite{13}, and (2) the myopic nature of RL's not exceptional long-horizon planning capacity that fails to account for long-term mission objectives \cite{14}. These limitations become particularly acute in the underwater tasks where AUVs must plan informative trajectories over\! extended time horizons while managing limited energy reserves \cite{27}.

\begin{figure*}[!t]
        \centering
        \includegraphics[width=0.818\linewidth]{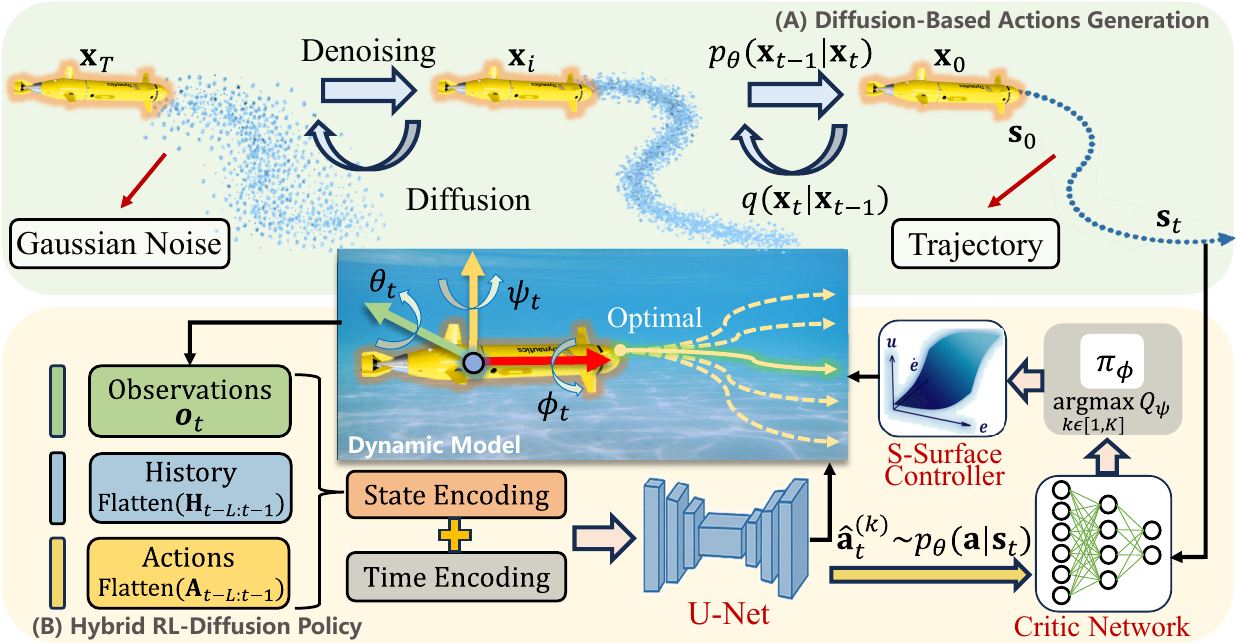}\vspace{-1.5mm}
        \caption{\small \textbf{Architecture of proposed framework for AUV robust control}. This framework consists of two components: (A) Diffusion-Based Actions Generation and (B) Hybrid RL-Diffusion Policy.}
        \label{fig_2}\vspace{-3mm}
\end{figure*}

Emerging as a powerful generative framework, diffusion model excels at capturing complex data distributions through iterative denoising \cite{15}. Unlike traditional approaches, it effectively models temporal state evolution—a critical feature for AUV control requiring smooth, physically realizable, and disturbance-resilient motions \cite{16}. Combined with RL, this synergy offers three advantages: (1) generating diverse, long-horizon trajectories that mitigate RL’s not exceptional planning capacity and help maintain stability under uncertain underwater dynamics \cite{17}; (2) enforcing physical constraints through denoising, ensuring feasible and robust control signals under hydrodynamic variability and disturbances \cite{18}; and (3) providing high-quality exploration trajectories that improve sample efficiency in sparse-reward settings and yield more resilient policies against real-world uncertainty \cite{19}. Overall, this fusion preserves RL’s adaptability while adding robustness and long-term planning, directly addressing challenges in AUV robust control.

 Based on the above analysis, this paper presents a diffusion-augmented RL framework for AUV robust control in underwater tasks, which can address these challenges through the following contributions:
\begin{itemize}
\item \textbf{Diffusion-Based AUV Actions Generation:}
We innovatively embed the diffusion model into RL for AUV robust control, generating physically realizable and high-quality actions. It employs a high-dimensional state encoding combining current observations with historical states/actions via a novel diffusion U-Net, enhancing long-horizon planning capacity in underwater tasks.

\item \textbf{Hybrid Learning Architecture:}  
Our framework integrates diffusion-guided exploration with RL policy optimization—the diffusion model generates diverse candidate actions, while the RL critic selects optimal ones for further execution. This co-adaptive approach improves exploration efficiency and AUV's policy stability in dynamic underwater environments.  

\item \textbf{Extensive Simulation Experiments:}  
The proposed framework shows superior robustness and flexibility over conventional approaches in the underwater 3D data collection task under challenging marine conditions.  
\end{itemize}

\section{METHODOLOGY}
In this section, we elaborate on the diffusion-augmented RL framework, beginning with the diffusion-based action generation mechanism and then extending to the hybrid RL-diffusion policy integrated with the S-surface controller.
\subsection{Diffusion-Based Actions Generation}
Building upon recent advances in generative modeling, we formulate the AUV robust control problem as a conditional denoising process that progressively refines noisy state sequences. Specifically, given a state sequence $\mathbf{x}_{0:T} = [\mathbf{s}_0,...,\mathbf{s}_T]$, the forward diffusion process gradually adds Gaussian noise via a variance schedule \cite{20}:\vspace{-1mm}
\begin{equation}
q(\mathbf{x}_t|\mathbf{x}_{t-1}) = \mathcal{N}(\mathbf{x}_t; \sqrt{1-\beta_t}\mathbf{x}_{t-1}, \beta_t\mathbf{I}),
\end{equation}
where $\beta_t$ follows a linear schedule from $10^{-4}$ to $0.02$ over $T=1000$ steps, thereby providing a smooth transition from the data distribution to isotropic Gaussian noise. 

In contrast, the reverse process learns to denoise through a U-Net architecture that incorporates both spatial and temporal information:\vspace{-1.5mm}
\begin{equation}
\epsilon_\theta(\mathbf{x}_t,t) = \text{U-Net}\Bigg(\text{Concat}\big[\text{MLP}(\mathbf{s}_0), \text{TimeEmb}(t)\big]\Bigg).
\end{equation}\vspace{-1.5mm}

To further ensure comprehensive state representation while maintaining computational efficiency, we design the following key components. These components are jointly responsible for enhancing the expressiveness of the diffusion process while controlling model complexity:
\begin{itemize}
\item \textbf{State Encoding}: First, the input $\mathbf{s}_t$ combines current observations with historical context:
\begin{equation}
\mathbf{s}_t = [\mathbf{o}_t, \text{Flatten}(\mathbf{H}_{t-L:t-1}), \text{Flatten}(\mathbf{A}_{t-L:t-1})],
\end{equation}
where $L=10$ represents the history length, chosen through ablation studies to balance memory requirements and predictive performance \cite{21}. In this formulation, the flatten operation preserves temporal correlations while reducing dimensionality.

\item \textbf{Time Embedding}: Moreover, to capture the diffusion process's progression, we employ sinusoidal embeddings with learned transformations:
\begin{equation}
\text{TimeEmb}(t) = \text{MLP}(\sin(10^4t/\tau)), \quad \tau=1000.
\end{equation}
This provides smooth, periodic representations that help the model effectively distinguish different noise levels.
\end{itemize}

\subsection{Hybrid RL-Diffusion Policy}
While the diffusion-based mechanism generates diverse candidate actions, it alone is insufficient to guarantee long-term optimality. To address this limitation, we further introduce a hybrid framework that integrates RL with diffusion-based proposals. As shown in Algorithm 1, the policy $\pi_\phi$ of our proposed framework integrates the exploratory capacity of the diffusion model with the decision accuracy of RL through a three-phase architecture.

\textbf{Diffusion Proposal Phase:}  
In the first phase, the model samples $K = 5$ diverse candidate actions from a denoising distribution:
\begin{equation}
\hat{\mathbf{a}}_t^{(k)} \sim p_\theta(\mathbf{a}|\mathbf{s}_t) = \mathcal{N}\big(\epsilon_\theta(\mathbf{s}_t,t), \sigma_t^2\big),
\end{equation}
where the noise scale $\sigma_t$ decreases over time, enabling broad exploration initially and finer refinement later.

\textbf{RL Selection Phase:}  
Building upon this proposal, an RL framework using TD3 for training based on a control-affine MDP $\mathcal{M} \triangleq (\mathcal{S}, \mathcal{A}, \mathcal{U}, C, f, g, d, \mathcal{R}_{\pi}, \gamma)$ guides the selection \cite{26, TD3}. Here, $\mathbf{s}_t$ evolves as:
\begin{equation}
\mathbf{s}_{t+1} = f(\mathbf{s}_t) + g(\mathbf{s}_t) C(\mathbf{a}_t) + d(\mathbf{s}_t),
\end{equation}
with control signal $\mathbf{a}_t$ drawn from $\pi(\mathbf{a}_t|\mathbf{s}_t)$, and $u_t = C(\mathbf{s}_t, \mathbf{a}_t)$. The terms $f$, $g$, and $d$ represent nominal dynamics and external disturbances (e.g., ocean waves), while $\mathcal{R}_{\pi}$ and $\gamma\in[0,1]$ denote the reward and discount factor, respectively.

Finally, a critic network evaluates the candidates and selects the action maximizing expected return:
\begin{equation}
\pi_\phi(\mathbf{s}_t) = \underset{k \in [1,K]}{\text{argmax}} \ Q_\psi(\mathbf{s}_t, \hat{\mathbf{a}}_t^{(k)}).
\end{equation}

In this way, the hierarchical framework balances diffusion's diversity with RL's long-term optimization, ensuring both robustness and adaptability in AUV's decision-making.

\textbf{S-Surface Controller Implementation}:  
To further translate the selected actions into reliable low-level control, we incorporate an S-Surface controller. This module provides robust tracking in dynamic underwater environments by combining nonlinear error compensation with disturbance rejection. The control law is given by \cite{22}:
\begin{equation}
u_t = \frac{2}{1 + \exp(-\zeta_1 e - \zeta_2 \dot{e})} - 1 + \Delta u,
\end{equation}
where $\zeta_1$ and $\zeta_2$ define the surface shape, and $\Delta u$ compensates for online disturbances. Consequently, this approach ensures finite-time convergence, smooth control signals, and adaptive gain adjustment based on error magnitude.

\begin{algorithm}[!t]
\caption{Diffusion-Augmented RL
Framework}
Initialize replay buffer $\mathcal{D}$ with diffusion-generated trajectories, TD3 critics and actor parameters.

\For{each episode $m = 1$ to $M$}{
Sample initial state $\mathbf{s}_0 \sim \mathcal{D}$.

\For{each time step $t$}{
    Generate candidate actions $\{\hat{\mathbf{a}}_t^{(k)}\}_{k=1}^K$.
    
    Select optimal action $\mathbf{a}_t = \pi_\phi(\mathbf{s}_t)$.
    
    Execute action $\mathbf{a}_t$ and observe $r_t$, $\mathbf{s}_{t+1}$.
    
    Store transition $(\mathbf{s}_t,\mathbf{a}_t,r_t,\mathbf{s}_{t+1})$ in $\mathcal{D}$.
    
    \While{In training phase}{
        Update diffusion model via reweighted evidence lower bound:
        \begin{equation}
        \nabla_\theta \|\epsilon - \epsilon_\theta(\sqrt{\alpha_t}\mathbf{s}_t + \sqrt{1-\alpha_t}\epsilon, t)\|^2
        \end{equation}
        with $\alpha_t = \prod_{s=1}^t (1-\beta_s)$.
        
        Update TD3 critics using clipped double Q-learning:
        
        \quad$y = r + \gamma \min_{i=1,2} Q_{\psi'_i}(\mathbf{s}',\pi_\phi(\mathbf{s}')+$
        \begin{equation}
        \text{clip}(\mathcal{N}(0,\sigma),-c,c))
        \end{equation}
        where $c=0.5$ prevents overestimation.
    }
}
}
\end{algorithm}
\vspace{-2mm}

\begin{table}[!t]
  \centering
  \caption{\small Parameters and hyperparameters configuration.} 
  \label{tab:hyperparameters}
  
  \begin{tabular}{lc}
    \toprule
    \textbf{Parameters} & \textbf{Values} \\
    \midrule
    Diffusion steps ($T$) & 1000 \\
    Denoising steps & 50
    \\
    Noise schedule ($\beta_t$) & Linear $1\times10^{-4}\rightarrow0.02$ \\
    History length ($L$) & 10 \\
    Candidate actions ($K$) & 5 \\
    U-Net hidden dimension & 256 \\
    Training batch size (RL/Diffusion) & 64 / 32 \\
    Hidden layer size (RL/Diffusion) & 128 / 256
    \\
    Learning rate (RL/Diffusion) & $1\times10^{-3}$ / $1\times10^{-4}$ \\
    Discount factor ($\gamma$) & 0.97 \\
            AUV maximum speed ($v_{\text{max}}$, $\omega_{\text{max}}$) & 2.3m/s, 0.26rad/s\\
        Propeller maximum revolution & 1525rpm \\
        Water density ($\rho$) & 1026$\text{kg/m}^{\text{3}}$ \\
        Control frequency & 20Hz \\
        Controller parameters for yaw ($\zeta_1,\zeta_2$) & 2, 1 \\
        Controller\! parameters for\! depth ($\zeta_1,\zeta_2$) & 1, 1 \\
    \bottomrule
  \end{tabular}\vspace{-3mm}
\end{table}

\section{EXPERIMENTS}
In this section, we perform extensive simulations to validate the effectiveness of the proposed framework.
\subsection{Task Description and Settings} 
The proposed framework is validated on a REMUS 100 AUV\! (1.6m,\! 31.9kg)\! with a three-part control system:
(1)\! a lightweight 2-layer 1D U-Net–style diffusion model with Group Normalization and Mish activation using a 256-dimensional embedding;
\!(2) a TD3-based policy with twin 2-layer MLP critics (128–128,\! ReLU) and a 3-layer actor (128 units/layer,\! tanh);\!
(3) an S-Surface controller at 20 Hz for real-time low-level control.
This framework jointly ensures stable action generation, policy optimization, and robust execution.

\begin{figure*}[!t]
        \centering
        \includegraphics[width=0.99\linewidth]{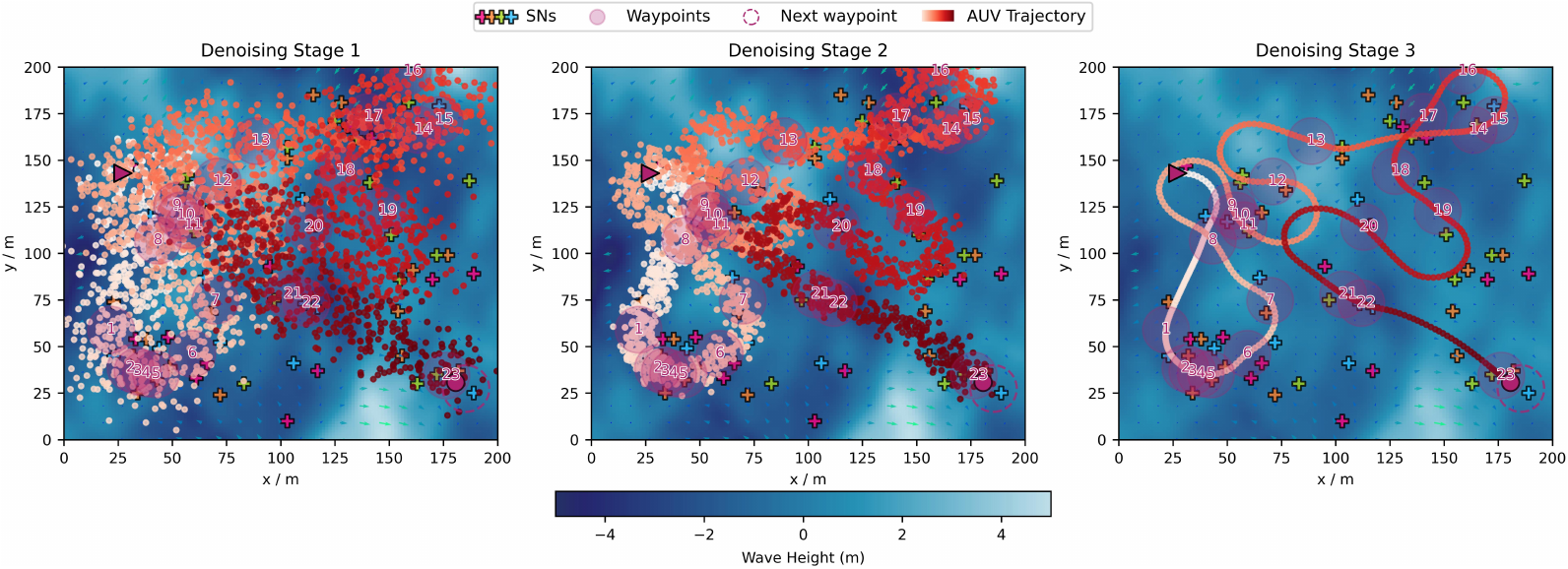}\vspace{-2mm}
        \caption{\small Visualization of five candidate actions across three denoising stages under the Diffusion+RL+S-Surface framework. Trajectories evolve from scattered exploration to smooth, task-aligned paths, demonstrating diffusion’s ability to generate diverse yet optimized plans.}
        \label{fig_6}\vspace{-3mm}
\end{figure*}

\begin{figure*}[!t]
        \centering
        \includegraphics[width=0.99\linewidth]{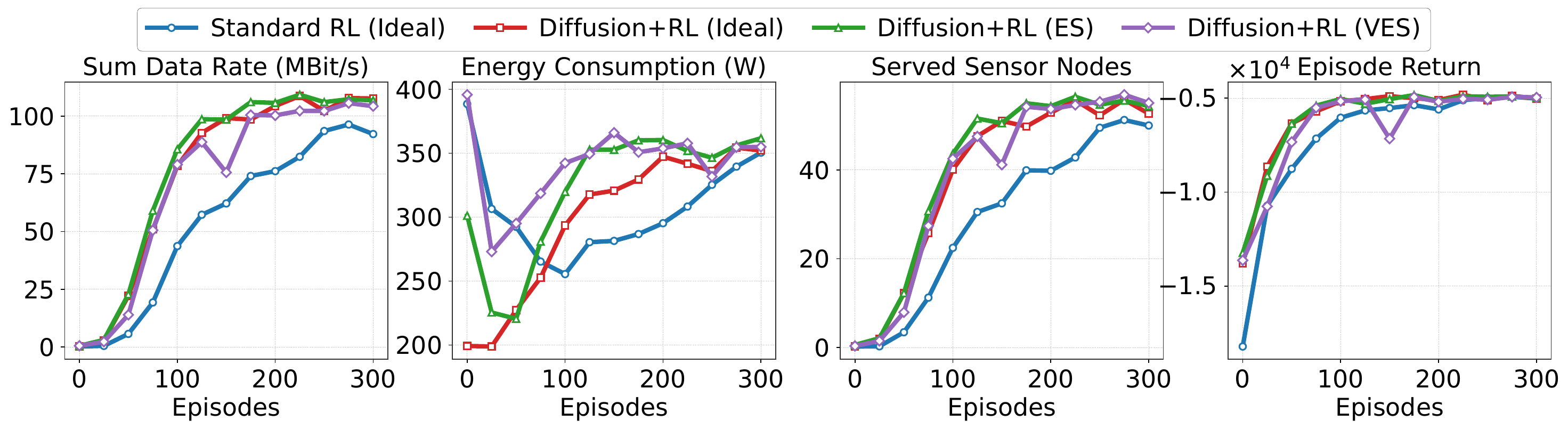}\vspace{-2mm}
        \caption{\small Performance comparison between diffusion+RL and standard RL across underwater data collection task metrics under ideal, ES and VES conditions, respectively.}
        \label{fig_3}\vspace{-3.5mm}
\end{figure*}

To comprehensively evaluate the performance of this framework, we adopt a 3D underwater data collection task in the absence of standardized benchmarks. 
In this scenario, multi-AUV and USV platforms collaborate to collect data from Sensor Nodes (SNs) within an Internet of Underwater Things (IoUT). 
The system jointly optimizes multiple objectives: maximizing Serviced SNs (SSN) and Sum Data Rate (SDR), while minimizing collisions and Energy Consumption (EC). 
Parameter details are summarized in TABLE I, with additional simulation references in \cite{30}.

Finally, simulations were conducted on a Ryzen~9 5950X CPU and RTX~3060 GPU using Python~3.12. 
The system completed 300 episodes of training in around 7 hours, reflecting reasonable computational efficiency.
    
\vspace{-1mm}  
\subsection{Experiment Results and Analysis}
We first illustrate the underlying mechanism of the diffusion model within the diffusion-augmented RL (Diffusion+RL) framework. As shown in Fig. 3, it visualizes the denoising process by executing five candidate actions generated at different reverse diffusion stages. Each subplot corresponds to one denoising step, where all five candidates are executed and visualized as colored trajectories. In the denoising stage 1, the trajectories are highly scattered, reflecting the exploratory nature of noisy action samples. As denoising progresses (middle), the trajectories begin to exhibit a more coherent structure and alignment with task-relevant waypoints. By the final stage, the candidates converge toward smooth, goal-directed paths that effectively balance coverage, feasibility, and environmental constraints. This progression—from noise-driven diversity to structured optimization—demonstrates how the diffusion model facilitates diverse action generation while progressively refining quality, enabling planning behaviors beyond the reach of conventional RL in sparse-reward underwater settings.

\begin{figure*}[!t]
    \centering
    \subfigure[\small Yaw and depth trajectories]{
        \includegraphics[width=0.464\linewidth]{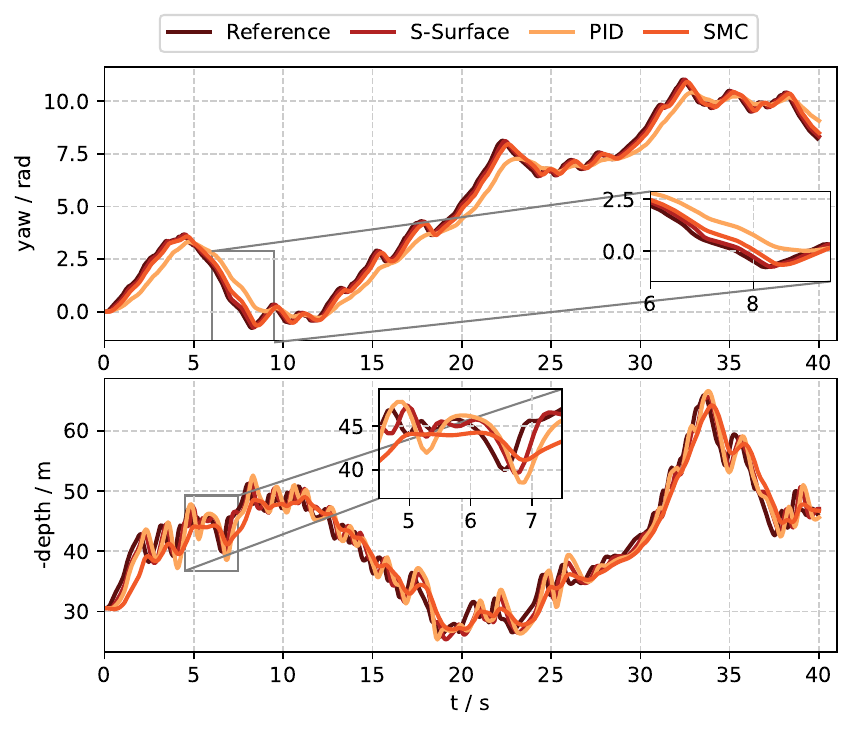}
        \label{fig_4a}
    }
    \subfigure[\small Yaw and depth tracking errors (MSE)]{
        \includegraphics[width=0.464\linewidth]{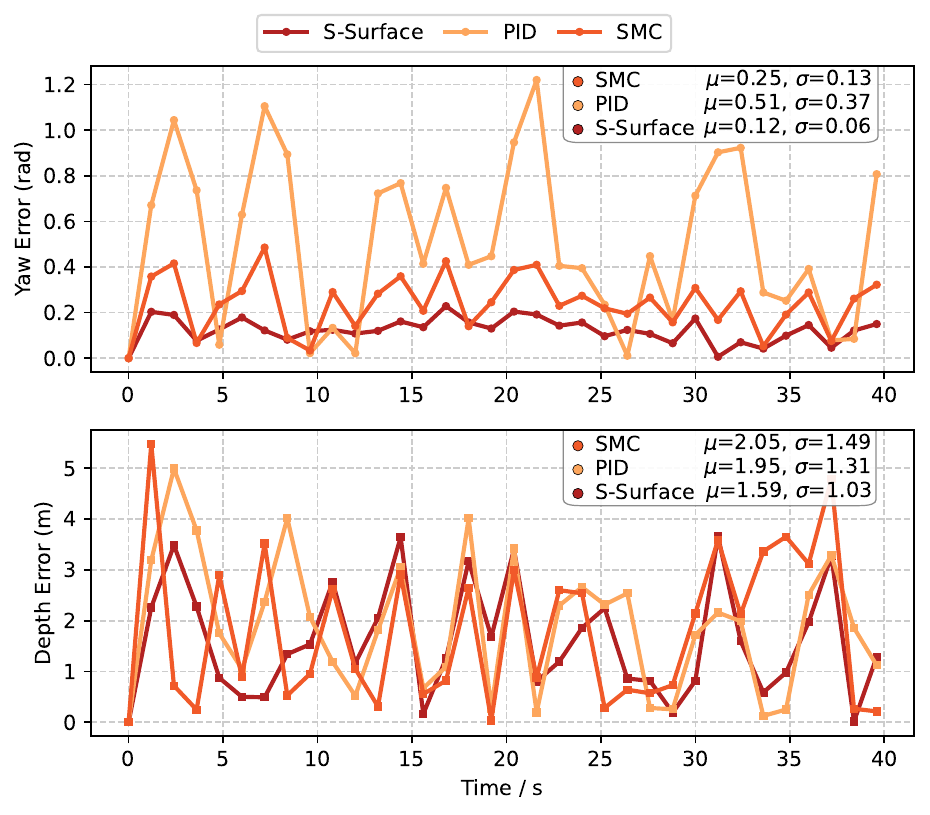}
        \label{fig_4b}
    }\vspace{-3mm}
    \caption{\small Tracking performance of Diffusion+RL with S-Surface, PID, and SMC controllers. S-Surface achieves the most accurate and stable tracking, with the lowest yaw and depth errors.
(a) Yaw and depth trajectories.
(b) Yaw and depth tracking errors (MSE).}
    \label{fig_4}\vspace{-3.5mm}
\end{figure*}

To evaluate the effectiveness of our Diffusion+RL framework, we compare its performance against a standard RL baseline under ideal, Extreme Sea conditions (ES) and Very Extreme Sea Conditions (VES) in the underwater data collection task, respectively. As illustrated in Fig. 4, Diffusion+RL exhibits significantly faster convergence, reaching stability within approximately 300 episodes—almost twice as fast as the standard RL counterpart—highlighting its enhanced sample efficiency. Moreover, by the end of training, Diffusion+RL consistently outperforms the baseline in terms of SDR and number of SSN, while maintaining similar overall EC. These results suggest that the integration of diffusion-based action generation not only expedites the learning process but also facilitates the development of effective control policies, enabling improved performance in the complex underwater environment.

\begin{figure*}[!t]
    \centering
    \subfigure[\small Yaw and depth trajectories]{
        \includegraphics[width=0.464\linewidth]{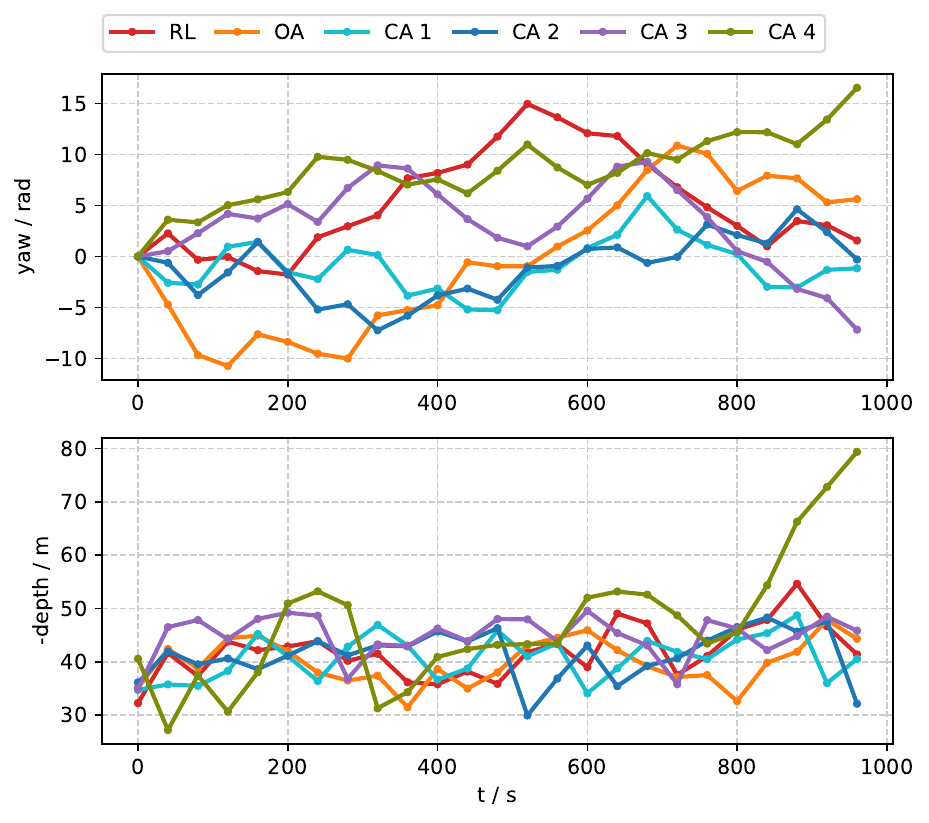}
        \label{fig_5a}
    }
    \subfigure[\small Q-value and task metrics]{
        \includegraphics[width=0.464\linewidth]{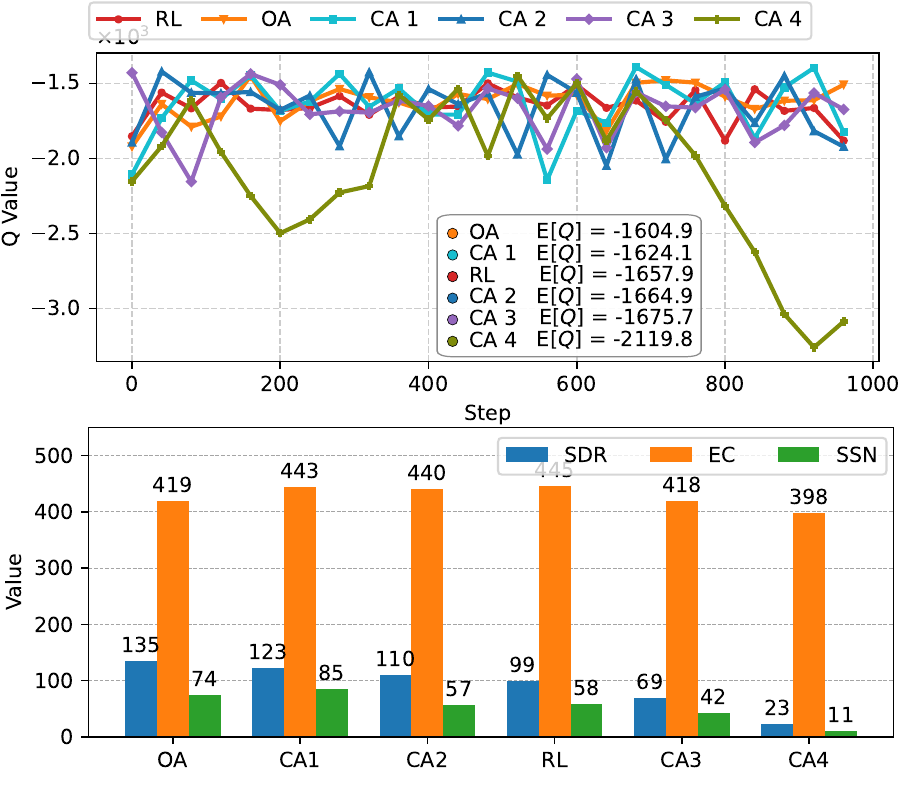}
        \label{fig_5b}
    }\vspace{-3mm}
    \caption{\small Comparison of six control strategies under the Diffusion+RL framework: the Q value-guided optimal action (OA), four diffusion-generated sub-optimal actions (CA1–CA4), and the RL+S-Surface baseline. (a) Yaw and depth trajectories. (b) Q-value and task metrics.}
    \label{fig_5}\vspace{-3mm}
\end{figure*}

Building upon these findings, we further investigate the influence of low-level controller design within the Diffusion+RL framework by comparing the performance of S-Surface, PID, and SMC controllers under the same high-level policy guidance. As shown in Fig. 5, the combination of Diffusion+RL with the S-Surface controller yields the most accurate trajectory tracking, exhibiting minimal yaw and depth deviations from the reference signals. Notably, the yaw and depth error plots on the right reveal that S-Surface maintains the lowest mean and standard deviation in both tracking errors (yaw: $\mu$=0.12, $\sigma$=0.06; depth: $\mu$=1.59, $\sigma$=1.03), reflecting high precision and temporal consistency. In contrast, the PID controller suffers from larger overshoots and higher error standard deviation, while the SMC controller exhibits pronounced chattering, which undermines control stability, particularly in dynamic segments. These results highlight the importance of nonlinear error shaping and disturbance rejection capabilities provided by S-Surface, which, when integrated with diffusion-guided action generation and RL, enable stable and robust tracking performance in challenging underwater conditions.

\begin{figure*}[!t]
        \centering
        \includegraphics[width=0.99\linewidth]{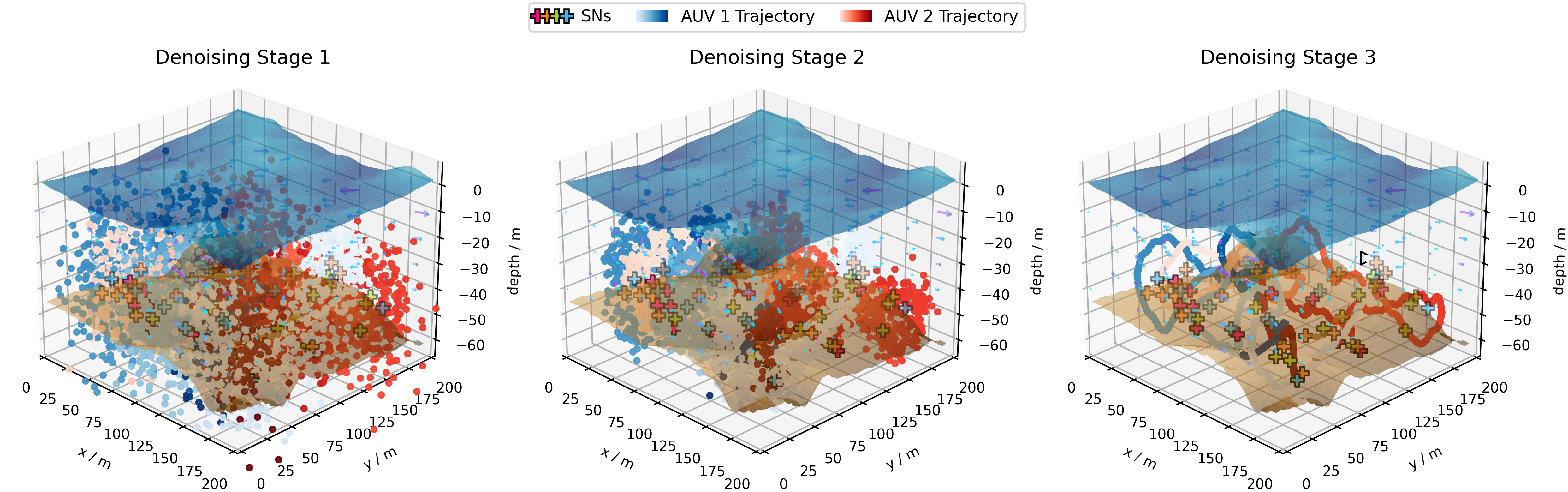}\vspace{-2mm}
        \caption{\small 3D trajectory evolution across denoising stages under Diffusion+RL+S-Surface. Candidates transition from scattered to coordinated paths, enabling robust multi-AUV planning and robust control.}
        \label{fig_7}\vspace{-3mm}
\end{figure*}

To further assess control accuracy under the Diffusion+RL framework, we compare six control strategies across one episode in the task: the Diffusion+RL+S-Surface controller using the RL's Q value-guided optimal action (OA), four sub-optimal candidate actions (CA1–CA4) generated by the diffusion model, and the standard RL+S-Surface controller without diffusion. As illustrated in Fig. 6(a), the OA configuration consistently maintains tighter yaw and depth trajectories with minimal deviations over time, while sub-optimal candidates—especially CA3 and CA4—exhibit significantly larger errors and instability. Although this plot is for qualitative illustration, this discrepancy is further reflected in the Q-value curves (Fig. 6(b)), where OA achieves the highest expected return (E[Q] = –1604.9), while CA4 performs the worst (E[Q] = –2119.8). Moreover, the bar chart shows that OA achieves the best overall balance, with a high SDR (135 Mbit/s), low EC (419 W), and a large number of SSN (74). In contrast, CA4 results in the lowest task quality and resource efficiency. Importantly, compared to the baseline using only RL, the Diffusion+RL framework—with diffusion-based candidate generation and Q value-guided action selection—yields significantly better control performance. This highlights the critical role of the diffusion model in producing diverse candidate actions and demonstrates that its integration with Q-value evaluation enables the AUV to execute more accurate and robust trajectories in complex underwater environments.

\begin{table}
  \centering
  \caption{\small Performance comparison of Diffusion+RL using different low-level controllers under ES and VES conditions.} 

    \begin{tabular}{ccccc}
      \toprule
      \multicolumn{2}{c}{Metrics} & \textbf{SDR} (MBit/s) $\uparrow$ & \textbf{EC} (W) $\downarrow$ & \textbf{SSN} (s) $\uparrow$\\
      \midrule
    \multirow{2}{*}{\textbf{S-Surface}} & ES & 105.1 $\pm$ 14.7 & 362.5 $\pm$ 22.8 & 55.4 $\pm$ 9.5 \\
    & VES &  101.0 $\pm$ 7.5 & 352.2 $\pm$ 6.5 & 54.4 $\pm$ 6.5 \\
    \midrule
    \multirow{2}{*}{\textbf{PID}} & ES & 88.6 $\pm$ 7.2 & 257.1 $\pm$ 17.6 & 44.5 $\pm$ 4.3 \\
    & VES & 86.1 $\pm$ 7.6 & 246.7 $\pm$ 20.5 & 43.6 $\pm$ 4.6 \\
    \midrule
    \multirow{2}{*}{\textbf{SMC}} & ES & 41.9 $\pm$ 7.0 & 225.7 $\pm$ 13.5 & 22.4 $\pm$ 3.1 \\
    & VES & 12.8 $\pm$ 5.2 & 173.0 $\pm$ 11.8 & 6.6 $\pm$ 2.7 \\
      \bottomrule 
  \end{tabular}\vspace{-4mm}
\end{table}

Furthermore, we also investigate the impact of low-level controllers within the Diffusion+RL framework. We evaluate three widely used control strategies—S-Surface, PID, and SMC—under two increasingly challenging marine environments: ES and VES conditions, as summarized in TABLE II. Among the compared methods, the S-Surface controller consistently delivers the best performance, achieving the highest SDR under both ES (105.1 ± 14.7 MBit/s) and VES (101.0 ± 7.5 MBit/s), while simultaneously serving the largest number of SSN. Although this comes at the cost of relatively higher EC, the overall task efficacy and trajectory informativeness justify the trade-off. The PID controller, while generally similarly performant, shows increased robustness under VES, with an SDR change from 88.6 ± 7.2 to 86.1 ± 7.6 MBit/s, albeit accompanied by increased energy expenditure. In contrast, the SMC controller exhibits the weakest adaptability to harsh marine conditions, with severe degradation in both SDR and SSN under VES (dropping to 12.8 ± 5.2 MBit/s and 6.6 ± 2.7, respectively), suggesting that it fails to maintain effective policy execution when subjected to compounded uncertainties. Taken together, the Diffusion+RL framework, when combined with the S-Surface controller, is able to achieve stable, reliable, and expected task performance in the complex and dynamic underwater environment.

Finally, Fig. 7 illustrates the evolution of five candidate actions at different denoising stages under the Diffusion+RL+S-Surface framework, providing an intuitive 3D view into the role of the diffusion model in the task. Each subplot corresponds to a denoising step in the reverse diffusion process, showing how AUV trajectories evolve in 3D space with respect to ocean surface, terrain, and sensor node distribution. In the initial stage (left), the five candidates exhibit wide spatial dispersion in both lateral and vertical dimensions, reflecting the model’s exploratory breadth. As denoising progresses (middle), trajectories begin to converge and align with task-relevant structures. By the final stage (right), the candidates have become smooth, structured paths that are both goal-directed and well-separated between the two AUVs. This transition from random dispersion to coordinated structure highlights the diffusion model’s ability to generate diverse yet high-quality control trajectories, supporting multi-AUV cooperation and task-aware planning and robust control in complex underwater environments.

    \vspace{-1mm}
    \section{CONCLUSIONS} 
 In this paper, we propose a diffusion-augmented RL framework for robust AUV control, addressing the challenges of nonlinear dynamics and uncertain underwater disturbances. The framework integrates: (1) a diffusion-based action generation mechanism with high-dimensional state encoding, enabling physically feasible actions and improved long-horizon planning, and (2) a hybrid learning architecture that combines diffusion-guided exploration with RL optimization, achieving sample-efficient and stable policy learning. Extensive simulations confirm superior robustness and flexibility compared to conventional control methods under dynamic marine conditions. Future work will focus on real-world experiments, further validating the adaptability of this framework for practical underwater operations.
    
    \bibliographystyle{IEEEtran}
    \bibliography{ICRA_lwy}
    \addtolength{\textheight}{-12cm}

\end{document}